\documentclass[sigconf, nonacm]{acmart}

\usepackage{amsmath}

\usepackage{amsfonts}

\usepackage{amssymb}

\usepackage{calrsfs}
\usepackage{dsfont}
\usepackage[greek,english]{babel}
\usepackage{url}
\usepackage{subcaption}
\usepackage{makecell}
\usepackage{xcolor}

\DeclareMathAlphabet{\pazocal}{OMS}{zplm}{m}{n}
\DeclareMathOperator*{\argmax}{argmax}
\newtheorem{definition}{Definition}

\usepackage{textcomp}
\renewcommand\footnotetextcopyrightpermission[1]{}
\begin{document}

\title{On Predictive Explanation of Data Anomalies}

\author{Nikolaos Myrtakis}
\email{myrtakis@csd.uoc.gr}
\affiliation{%
  \institution{University of Crete}
  \city{Heraklion}
  \country{Greece}
}

\author{Ioannis Tsamardinos}
\email{tsamard.it@gmail.com}
\affiliation{%
  \institution{University of Crete}
  \city{Heraklion}
  \country{Greece}
}

\author{Vassilis Christophides}
\email{vassilis.christophides@ensea.fr}
\affiliation{%
  \institution{ENSEA}
  \city{Cergy}
  \country{France}
}

\renewcommand{\shortauthors}{Myrtakis et al.}

\begin{abstract}
  Numerous algorithms have been proposed for detecting anomalies (outliers, novelties) in an unsupervised manner. Unfortunately, it is not trivial, in general, to understand why a given sample (record) is labelled as an anomaly and thus diagnose its root causes. We propose the following {\em reduced-dimensionality, surrogate model approach} to explain detector decisions: approximate the detection model with another one that employs only a small subset of features. Subsequently, samples can be visualized in this low-dimensionality space for human understanding. To this end, we develop {\bf PROTEUS}, an AutoML pipeline to produce the surrogate model, specifically designed for feature selection on imbalanced datasets. The PROTEUS surrogate model can not only explain the training data, but also the out-of-sample (unseen) data. In other words, PROTEUS produces {\bf predictive} explanations by approximating the decision surface of an unsupervised detector. PROTEUS is designed to return an accurate estimate of out-of-sample predictive performance to serve as a metric of the quality of the approximation. Computational experiments confirm the efficacy of PROTEUS to produce predictive explanations for different families of  detectors and to reliably estimate their predictive performance in unseen data. Unlike several ad-hoc feature importance methods, PROTEUS is robust to high-dimensional data. 
\end{abstract}


\maketitle

\section{Introduction}
\label{sec:introduction}

Detection of ``anomalous'' samples (records, instances), called \emph{anomaly detection}, is an important problem in machine learning. It is conceptually related to outlier and novelty detection in several application settings. The anomalous samples may indicate mislabelled data, catastrophic measurements or data entry errors, bugs in data wrangling and preprocessing software, or other interesting phenomena. 

Numerous \textbf{unsupervised} algorithms (e.g., IF \cite{Liu2008IsolationF}, LOF \cite{Breunig2000LOFID}, LODA \cite{Pevn2015LodaLO}) to detect anomalies (hereafter {\bf detectors}) have been proposed. The most advanced ones detect anomalies in a multi-dimensional fashion, simultaneously considering all feature values. Unfortunately, detectors, in general, do not explain why a sample was considered as abnormal, leaving human analysts with no guidance about their root causes, insight to take corrective actions, or remedy their effect.

Several methods for {\bf explaining anomalies} have been proposed, hereafter {\bf explainers}. {\em The explanations often take the form of a subset of features} called a {\bf subspace} in the literature. The idea is that {\em by examining only the explaining features suffices to determine whether the sample is an anomaly or not according to the detector}. 

Existing methods can be categorized to those that provide {\bf local explanations} (point-based) that pertain to a single sample, or {\bf global explanations} (a.k.a. set-based) to simultaneously explain all training samples. The latter is important in order to reduce the burden of human analysts to inspect possibly different explanations for each anomaly. We should stress that global explanation is different from clustering as the former's objective is to provide a subspace segregating the anomalous from normal samples. Explainers may be {\bf specific} to a detection algorithm or detector-{\bf agnostic}, hence applicable post-hoc to any detection algorithm. As reported by several independent experimental studies, e.g. \cite{Goldstein2016}, there is no detector outperforming all others on all possible datasets. Hence, researchers cannot just design a specific explainer for the optimal detector; it may thus be preferable to design optimal agnostic explainers. Explainers may also be categorized as {\bf descriptive} in the sense that they explain the samples used to train the detector. Explainers that return explanations that  generalize to unseen data are {\bf predictive} ones. The importance of predictive explanations has been recognised in Explainable AI to avoid recomputing explanations on every new batch of data. 

\begin{figure}[ht]
    \centering
    \includegraphics[width=1.0\linewidth]{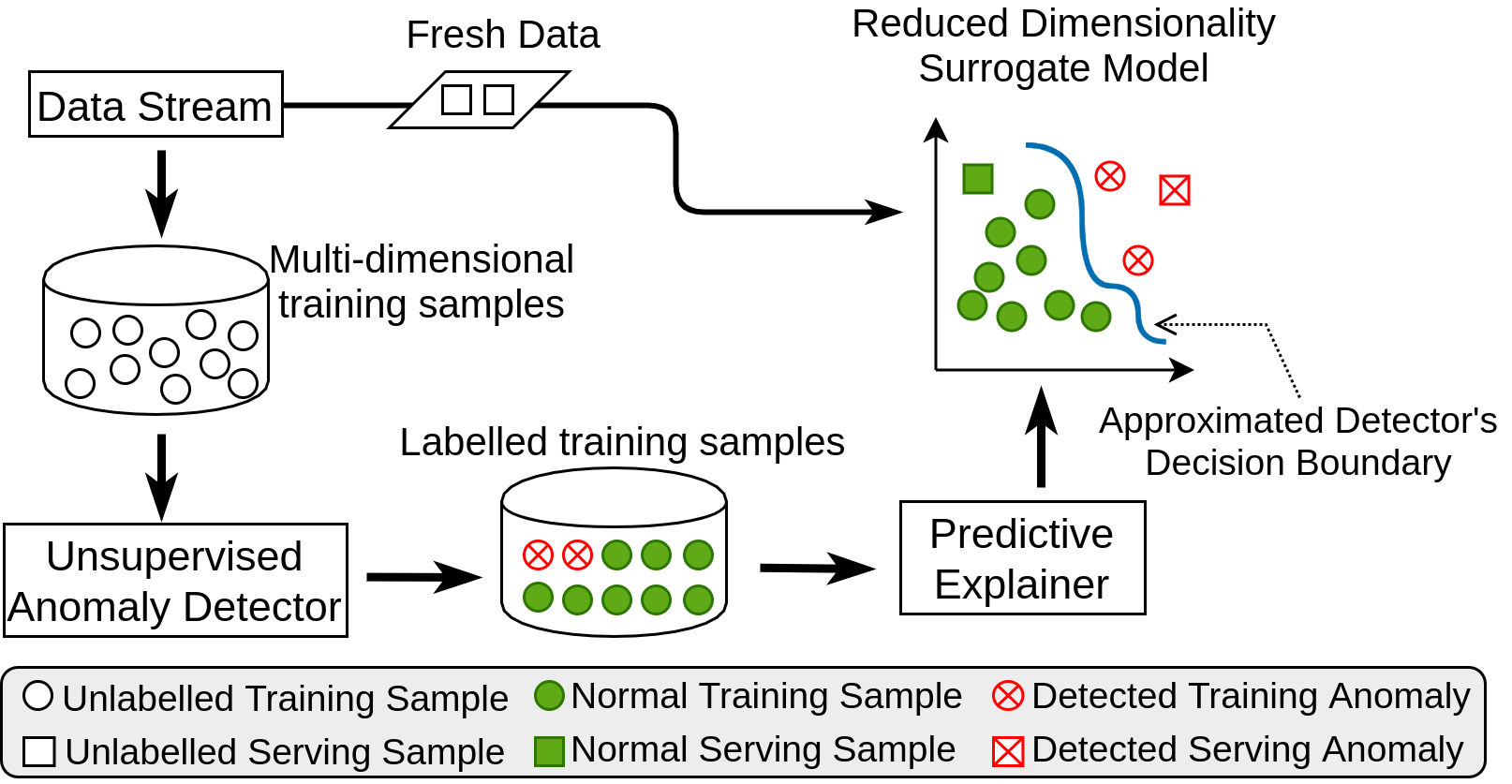}
    \caption{Predictive Anomaly Explanation Pipeline}
    \label{fig:predPipeline}
\end{figure}

Figure \ref{fig:predPipeline} illustrates how predictive explanations can be used in data validation pipelines monitoring the data fed to downstream ML models. Given that in real application settings it is difficult or even impossible to label data as anomalous or normal \cite{Goldstein2016}, unsupervised detectors are initially used to spot anomalies. Then, a predictive anomaly explainer could be used by human analysts to reveal the root causes of the detected anomalies and decide subsequent corrective actions. It is essentially a surrogate model, trained with a small subset of the original features that serve as explaining feature subspace. Depending on the quality of the approximation of the decision boundary of an unsupervised detector, the surrogate model can be also used to detect anomalies in fresh data, i.e., new batches of data, by completely bypassing the need to rerun the detector.

In this paper, we propose {\bf a novel method to produce global, predictive explanations} called {\bf PROTEUS}\footnote{Proteus or \textgreek{Πρωτεύς} in Greek, means `first' and is a minor sea God and son of Poseidon.}. PROTEUS is \emph{detector agnostic}, and can be used to approximate the decision boundary of any detector. We should stress that prior work on detector agnostic explainers like CA-Lasso \cite{MicenkovaNDA13} and SHAP \cite{LundbergL17} but also detector specific explainers like LODA \cite{Pevn2015LodaLO} produce explanations that are only {\bf local and descriptive}.

PROTEUS essentially constructs a {\em reduced-dimensionality, surrogate model} that approximates the behavior of a detector with fewer features. Since the detector is labelling the samples as anomalies or not, the problem of finding such a model reduces to a {\em supervised predictive modeling with feature selection} problem. In order for the surrogate model to also explain unseen samples, it has to approximate the detector's decision boundary and not simply interpolate the anomalies (overfit) in the training data. To this respect, {\em the quality of approximation should be estimated using out-of-sample performance estimation protocols} like $K$-fold cross validation (CV). To build the model, any combination of feature selection algorithm with a classifier could be employed. However, ideally one should optimize the combination of algorithms and their hyper-parameter values to achieve the best approximation with the samples at hand.

The above requirements for tuning and estimating  generalization performance of predictive models are nowadays addressed in the {automated machine learning} ({\bf AutoML}) systems \cite{Hutter2018}. In this respect, {\bf producing predictive anomaly explanations can be solved as an AutoML problem}. Unfortunately, the majority of existing tools such as auto-sklearn do not perform feature selection. In addition, they do not exploit the fact that the data can be augmented with new samples (pseudo-samples) that can be labelled by the detector, to improve performance. Finally, their performance estimates are often overestimated\footnote{https://www.kdnuggets.com/2020/12/trust-automl.html}, particularly for imbalanced datasets. To address the above issues, PROTEUS makes the following contributions: 

(1) In Section \ref{sec:problemdef}, we introduce a novel AutoML engine specifically designed to support feature selection and classification on imbalanced datasets. Unlike existing explainers, PROTEUS outputs not only a {\em small-sized feature subset serving as explanation} but also a {\em surrogate model fitted on this subset} to explain unseen samples, as well as a {\em reliable out-of-sample (predictive) performance estimation}.

(2) To produce such output, PROTEUS AutoML relies on {\em advanced design choices} described in Section \ref{sec:proteusPipeline}, such as supervised oversampling, group-based stratification, and a special variant of Cross-Validation with Bootstrap Bias Correction ({\bf BBC}) \cite{TsamardinosGB18}.

(3) Thorough computational experiments presented in Section \ref{sec:experiments} we show the efficacy and robustness of PROTEUS in synthetic and real datasets of increasing dimensionality. Last but not least,  our experiments show that PROTEUS approximates accurately the performance of a specific explainer (LODA) in a detector-agnostic fashion.

(4) We formally define in Section \ref{sec:problemdef} descriptive and predictive explanations, originally introduced in our work.

(5) We assess in Section \ref{sec:qualityAssessment} the merit of the idea to use PROTEUS to correct the decisions of the unsupervised anomaly detectors. Specifically, we study the disagreements of classification to anomalies between the surrogate PROTEUS model and the detector. We show that PROTEUS can often correct the false positives of false negatives as identified by the detector.  

(6) We propose a new visualization method for presenting the global explanations found by PROTEUS as spider charts. The visualizations provide insight regarding the combination of feature values that lead to calling a sample as anomalous or not. 

(7) We position in Section \ref{sec:relWork} PROTEUS w.r.t. various categories of related work on explaining anomalies in unsupervised and supervised settings.

Finally, in Section \ref{sec:conclusion} we conclude the paper and discuss directions of future research.





\section{Problem Definition}
\label{sec:problemdef}

In this section, we formalize the notion of descriptive explanations inspired by \cite{Gupta2018BeyondOD} and we introduce the novel concept of predictive explanations. 

Let $D = \{x_1, \ldots, x_n\}$ be a dataset of $n$ samples, where each sample $x \in \mathbb{R}^d$. An {\bf Anomaly Detector} $A$ is essentially a function that scores the "anomalousness" of samples in $D$ according to an unsupervised \textbf{Anomaly Model}: $\omega_A: \mathbb{R}^d \rightarrow \mathbb{R}$. Continuous scores are then converted into dichotomous decisions using a threshold choice method \cite{YangRF19}. Given a threshold $T$ and sample $x \in D$, a {\bf binary} Anomaly detector is a function $\omega'_A: \mathbb{R} \rightarrow \{0, 1\}$ defined as follows: $\omega'_A(x) = \mathds{1}[\omega_A(x) > T]$. The value $\omega'_A(x) = 1$, semantically denotes the identification of an anomaly.

\begin{definition}
The descriptive explanation $\mathcal{D}$ of a set of anomalies $O = \{x ~|~ \omega'_A(x) = 1, x \in D\}$, is a subset of features $S$, where $ |S| = b \ll d$, that maximizes the {\bf cumulative score} for a set of anomalies:
\begin{align}
    \label{eq:descexplanation}
    \begin{split}
        \mathcal{D} = \argmax_{S} ~ & \sum_{x \in O} \omega_A (x[S]) \\
        & \textnormal{s.t.} ~~ |S|=b
    \end{split}
\end{align}
\noindent
where $[\cdot]$ denotes the projection of $x$ over the features in $S$ composing its explanation. 
\end{definition}

\noindent
Such explanations are called \emph{descriptive} as they are computed for every new batch of anomalous and normal samples. In order to make explanations also \emph{discriminative} for unseen data, we need to consider \emph{predictive} explanations i.e., a hyperplane of reduced dimensionality that separates the anomalies from the normal samples when training a classifier over the output of an unsupervised anomaly detector. To produce explaining hyperplanes we need to evaluate alternative surrogate models built using different classification algorithms $h \in H$, where $h$ is fitted in a lower dimensional space, produced in turn by different feature selection algorithms $g \in G$ that consider the labels returned by different anomaly detectors.   

In a nutshell, \emph{predictive} explanations are produced by solving an AutoML problem \cite{Hutter2018}. We denote the combination of an algorithm $g$ and $h$ with their respective hyper-parameter values $a$ and $b$ as a configuration $\theta$, which is a function $f = h(g(\cdot, a), b)$. The function $f$ first applies the specified feature selection algorithm $g$ with hyper-parameters $a$ to some input data and the result is then used to train a classifier $h$ with hyper-parameters $b$. 
Let $D^l = \{(x_1, \hat{y}_1), ..., (x_n, \hat{y}_n)\}$ be a the augmented dataset $D$ enriched with the anomaly labels as indicated by the detector model: $\hat{y_i} = \omega'_A(x_i)$. 

\begin{definition}
\label{def:predExplanation}
The predictive explanation $\mathcal{P}$ is the hyperplane that comprises of a minimal subset of features $S$ leading to an optimal surrogate model $h$ w.r.t. a performance metric $Q$:
\begin{align*}
    \mathcal{P} = \argmax_{S} \max_{h} Q(h(D^l[S]))
\end{align*}
\end{definition}

\noindent
Given the dataset $D^l$, the objective is to build a reduced-dimensionality surrogate model $f$ trained with some data $D^l_{\mathit{train}}$ to \emph{best approximate the detector's decision boundary}. To assess the quality of the approximation, $f$ has to generalize to unseen data $D^l_\mathit{test}$ which were not used during the training of $f$. Therefore, the objective is to find the configuration $\theta^*$ that contains the tuple $\langle h^*, g^*, a^*, b^* \rangle$ maximizing a performance metric $Q$:
\begin{align}
    \label{eq:bestConf}
    \theta^* = \argmax_{\theta} ~ Q(f(D^l_{\mathit{train}},  \theta), ~ D^l_{\mathit{test}})
\end{align}

\noindent
The last step is to train the best configuration using all available data, to produce the final surrogate model $f(D^l, \theta^*)$ i.e., a  model $h^*(D^l[S], b^*)$ that is used to predict the "anomaloussness" of unseen samples using only a subset of features $S = g^*(D^l, a^*)$.

As anomalies are rare, the quality of performance of a predictive explanation requires evaluation metrics that are insensitive to the class distribution. In this respect, PROTEUS relies on optimizing the area under the Receiver Operating Characteristic (ROC AUC) curve (hereafter \textbf{AUC}). Given a minimal subset of features and a classifier, \emph{AUC equals the probability that the classifier scored higher an anomalous than a normal sample}. Discovering such minimal subset is a challenging task as the search space is exponential and features in the input dataset may be both \emph{irrelevant} or \emph{redundant} w.r.t. to the predictive outcome. PROTEUS relies on effective and efficient \emph{feature selection} algorithms \cite{Lagani2016FeatureSW, TsamardinosBKPC19, Tibshirani1996RegressionSA} to extract predictive explanations in a supervised setting. 


\section{Producing Global, Predictive Explanations with PROTEUS}
\label{sec:proteusPipeline}

\begin{figure*}
    \centering
    \includegraphics[width=1.0\linewidth]{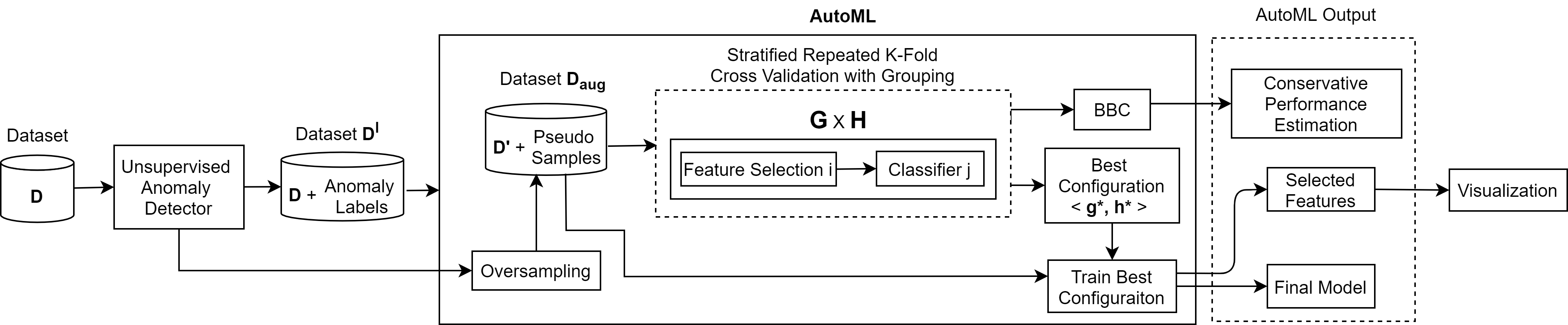}
    \caption{Proteus AutoML Pipeline for Anomaly Detection and Explanation}
    \label{fig:proteusPipeline}
\end{figure*}

Figure \ref{fig:proteusPipeline} illustrates the main steps of the pipelines automatically generated by PROTEUS. We proceed with explaining each step as well as the underlying design choices.

\noindent{\bf Producing Predictive Explanations as a Supervised Task}. First, the anomaly detector runs in dataset $D$ for producing the anomaly scores which are then transformed into binary labels (anomaly or not) in dataset $D^l$. Producing a surrogate model of lower dimensionality becomes a supervised, binary classification task with feature selection, where the outcome is the label of the unsupervised detector. We note that {\em data are standardized} for subsequent steps. 

\noindent{\bf Oversampling}. $D^l$ is expected to be highly imbalanced (w.r.t. the outcome), as anomalies are rare. Imbalanced datasets are statistically challenging for any ML classifier. One technique to alleviate the problem is {\em oversampling} the minority class. We focus on {\em synthetic minority oversampling}, i.e., the samples are perturbed by adding noise to the values of the features, creating new samples called {\em pseudo-samples}. In common (unsupervised) oversampling methods, for small enough perturbations the pseudo-samples are {\em assumed} to remain in the minority class. An assumption that strongly depends on the definition of what is considered ``small-enough''. However, one can take advantage of the detector model produced in the first step is available to query regarding the label of a pseudo-sample. In other words, PROTEUS oversampling is {\em supervised} as in case of explanation methods for black-box predictive models \cite{Ribeiro0G16}. Intuitively, oversampling probes the region around the anomalies and perturbs these samples to examine if they cross the detector's decision boundary or not. It thus effectively increases the available sample size for the classification, potentially increasing the quality of the approximation with the surrogate model. For each anomalous sample $a$ it produces $ps$ pseudo-samples per anomaly by adding a perturbation vector $p$ to $a$: $a' \leftarrow a + p$. Each $p$ follows a multi-variate ($d$-dimensional) normal distribution with zero mean and an isotropic, diagonal, covariance matrix $\sigma I$; $\sigma$ is a hyper-parameter of the algorithm which we set to 0.1 for all the computational experiments. If $a'$ is labelled as an anomaly it is appended to the oversampled dataset $D_{aug}$, otherwise, another pseudo-sample is produced. 

\noindent{\bf Hyper-Parameter Optimization Space}. To produce small-sized explanations, PROTEUS relies on feature selection algorithms, while to produce the surrogate model, a classifier is required. Most classification algorithms also accept a set of hyper-parameter values that also need to be tuned. We will call a combination of feature selection and classification algorithms and their hyper-parameters values as a {\em configuration}. {\em Each configuration is a pipeline that accepts a dataset and produces a classification model and corresponding selected features}. PROTEUS searches the configuration space for the one that leads to an optimal model by performing a simple grid search. This is, the search space of configurations is formed by the Cartesian product $\pazocal{G} \times \pazocal{H}$ (see Figure \ref{fig:proteusPipeline}) where $\pazocal{G}$ ($\pazocal{H}$, respectively) is the set of all feature selection (classification) algorithms with bounded hyper-parameter values. As our choices for feature selection algorithms, we include the Statistical Equivalent Signatures (SES) \cite{Lagani2016FeatureSW}, Forward-Backward with Early Dropping (FBED) \cite{TsamardinosBKPC19}, and Lasso. All of them guarantee to return the optimal feature subset (Markov Blanket in Bayesian Networks) under certain broad (but different for each algorithm) conditions, removing not only \emph{irrelevant}, but also \emph{redundant} features. In general, SES and FBED tend to return smaller feature subsets than Lasso, with a small drop in predictive performance \cite{TsamardinosBKPC19}.

Moreover, as anomaly explanation targets human analysts, {\em we limit the number of features selected up to 10}, ranking them based on their score given by the corresponding algorithm (e.g. Lasso coefficients). We selected linear as well as non-linear classifiers considering two facts (a) the extensive experimental results of \cite{DelgadoCBA14}, (b) the fact that deep neural network architectures are almost certain to overfit in very low sample sizes, both in terms of total sample size and the size of the rare class. The present selection of classifiers comprises of: (i) Support Vector Machines, (ii) Random Forest and (iii) K-Nearest-Neighbors. Due to space constraints, we report the hyper-parameters in our GitHub repository\footnote{\url{https://git.io/JtCwU}}. Finally, the number of pseudo-samples to create per anomaly, called $ps$ is also tuned as a hyper-parameter taking values in $\{0, 3, 10\}$. Of course, additional classifiers and feature selection algorithms can be easily integrated in PROTEUS. In total, PROTEUS tried 1800 configurations.

\noindent{\bf Estimating Performance for Tuning}. What is considered as the optimal configuration, out of all tried, is {\em the one that leads to models with the highest expected out-of-sample (unseen samples) predictive performance}. It is important to estimate this quantity accurately, i.e., with small variance. {\em A smaller variance of estimation increases the probability that the truly optimal configuration will be selected, and thus improves the quality of the final model}. Estimation is challenging when there are only few anomalies in the dataset. Indicatively, the synthetic dataset used in our experiments (see Section \ref{sec:datasets}) contains 10 anomalies out of 867 samples. 

To estimate the expected out-of-sample performance, PROTEUS employs a {\em Stratified, $R$-Repeated $K$-fold Cross Validation with Grouping} protocol. We now explain each part of the protocol. We assume that the reader is familiar with the Standard $K$-fold Cross Validation ({\bf CV}, hereafter).
The {\em Stratified CV} is a variant where the partitioning to folds is performed under the constraint that the distribution of the classes in each fold is approximately the same as the one in the full dataset \cite{TsamardinosGB18}. Stratification reduces the variance of estimation for imbalanced data and classes with very few samples ({\em ibid}). To further reduce the variance of estimation we repeat the CV process multiple times $R$ and take the average ({\em $R$-Repeated CV}). Multiple repeats reduce the variance component due to the stochasticity of the specific partitioning. Prior work has shown its benefits {\em ibid}. Finally, we come to {\em Grouping}. By CV with Grouping we indicate a variant of CV that handles grouped samples (a.k.a. as clustered samples in statistics, not to be confused with clustering of samples). These are samples that are not independently sampled and maybe correlated given the data distribution. Such samples are repeated measurements on the same subject, as an example. In our context, {\em an anomaly and its pseudo-samples are grouped}: information from a pseudo-sample in the training set {\em leaks} to predicting the corresponding anomaly in the test fold. To avoid information leakage, CV with grouping partitions to folds with the constraint that all samples of a group remain in the same fold. In our experiments, we set the number of folds $K=10$ and the repeats $R=5$. Hence, each application of the current version of PROTEUS trains $(K \cdot R \cdot \textnormal{\# Configurations} + 1) \cdot ps = 90,003$  models.

\noindent{\bf Producing the Final Surrogate Model and Feature Subset}. 
The final model is trained using all available samples (the full $D_{aug}$) with the best configuration found, denoted with $\langle F^*, C^* \rangle$ in Figure \ref{fig:proteusPipeline}. This configuration also produces the final subset selection (anomaly explanation). The reasoning is that most algorithms (and hence, configurations) are expected to produce better quality models and improved feature selection with more available sample. The models trained during the CV are only employed for selecting the optimal configuration and providing estimates. 

\noindent{\bf Estimating the Out-of-Sample Performance}.
We now consider how the performance estimate of the final model is produced. Let us assume that 1000 configurations are tried and the best found has a CV estimate of 0.90 AUC. Unfortunately, {\em the CV estimate of the best configuration is optimistic and should not be returned}, i.e., the actual AUC is expected to be lower. The reason is that our estimate is the best out of 1000 tries \cite{TsamardinosBKPC19,JensenC00}.
The phenomenon is conceptually similar to the multiple hypothesis testing problem in statistics. 
In small sample sizes, the over-optimism is particularly striking.
In this respect, we  apply the Bootstrap Bias Correction ({\bf BBC}, hereafter) to our CV estimates \cite{TsamardinosGB18} that corrects for this optimism. {\em This leads to returning conservative estimates of performance on average}.


\section{Experimental Evaluation}
\label{sec:experiments}

PROTEUS was implemented in Python 3.6 and evaluated on several synthetic and real-world datasets as described below. The code and the datasets used in our experiments are available in our GitHub repository. All experiments were performed in a Linux Desktop computer with a 4-core Intel i5 processor and 32GB of memory.


\subsection{Synthetic and Real Datasets}
\label{sec:datasets}

We focus on datasets where the samples are \emph{independent and identically distributed (i.i.d.)} and contain numerical features. We employ a \emph{synthetic} dataset, where anomalies have been simulated so that a minimal, global, predictive explanation (feature subset) is both achievable and known. The presence of this gold-standard allows us to evaluate how well PROTEUS identifies it. Specifically, we selected randomly one of the 100-dimensional datasets introduced in \cite{KellerMB12}. Some anomalies have been generated in a way that makes them outliers according to a subset of 2 of these features, call it $S_{2d}$, and some according to a subset with 3 (other) features, call it $S_{3d}$. Thus, the subset of these 5 features $S = S_{2d} \cup S_{3d}$ forms the \emph{gold-standard of global explanation for all anomalies}. On this \emph{parent} synthetic dataset, we added irrelevant features with randomly selected values following a normal distribution with zero mean and standard deviation of one. We ended up with 5 synthetic datasets having 20, 40, 60, 80 and 100 dimensions. All of them contain 867 samples with 10 anomalies i.e., the anomaly ratio is $\approx$ 1\%. Such datasets have been frequently used in the literature of anomaly explanation \cite{MicenkovaNDA13, DangANZS14, KellerMWB13, VinhCBLRP15}, because: (a) the features in an explaining subspace (e.g, $S_{2d}$) are correlated so feature cannot be selected independently; (b) anomalies are recognized as such either in $S_{2d}$ or $S_{3d}$, but in no other strict subset. Thus, only multivariate detection algorithms and corresponding models will achieve high performance. Hence, PROTEUS must approximate a potentially more complex model.


We additionally consider \emph{real-world datasets} that are widely-used in the evaluation of anomaly detectors. Specifically, we selected the Wisconsin-Breast Cancer, Ionosphere and Arrhythmia, originally from the UCI Machine Learning repository, as defined for anomaly detection purposes in Outlier Detection DataSets (ODDS) repository\footnote{http://odds.cs.stonybrook.edu/}. They were chosen to ensure that the detectors employed achieve reasonable performance, and thus explanation makes sense. The dataset characteristics and detector performances are shown in Table \ref{tab:realDataSpecs}. Wisconsin-Breast Cancer and Ionosphere contain two classes. The minority classes in both datasets are considered as anomalies. For Arrhythmia, eight sub-classes were merged to form the anomaly class. Finally, we added irrelevant features following the procedure described in synthetic datasets constructing three additional datasets per real-world dataset with 30\%, 60\% and 90\% irrelevant feature ratio.


\begin{table}
    \fontsize{9.0}{10.0}
    \selectfont
    \centering
    \caption{Characteristics of datasets and AUC performance of detectors during training. We denote the parent synthetic dataset as P. Synthetic, the number of features and samples as \#F and \#S and the anomaly ratio as A.R.}
    \begin{tabular}{lccc||ccc}
    \hline
        Dat. Name & \textbf{\#F} & \textbf{\#S} & \textbf{A.R.} & \textbf{IF} & \textbf{LOF} & \textbf{LODA} \\ \hline
        P. Synthetic & 5 & 867 & 1\% & 0.96 & 1.0 & 0.92 \\
        W. Br. Cancer & 30 & 377 & 5\% & 0.95 & 0.94 & 0.96\\
        Ionosphere & 33 & 358 & 36\% & 0.85 & 0.93 & 0.87   \\
        Arrhythmia & 257 & 452 & 15\% & 0.80 & 0.74 & 0.75 \\
    \hline
    \end{tabular}
    \label{tab:realDataSpecs}
\end{table}

\subsection{Experimental Setting}
In our experiments, we selected three widely-used unsupervised anomaly detectors that employ different anomalousness criteria, namely Local Outlier Factor (LOF) \cite{Breunig2000LOFID} as a representative of \emph{density-based}, Isolation Forest (IF) \cite{Liu2008IsolationF} as a representative of \emph{isolation-based} and Lightweight On-line Detector of Anomalies (LODA) \cite{Pevn2015LodaLO} as a representative of \emph{projection-based} detectors. Regarding the hyper-parameters, for IF we used 100 trees and 256 sub-sample size, for LOF we used $K = 15$ and for LODA we used 100 projection vectors. To assess the predictive power of a surrogate model produced by PROTEUS we stratified and splitted each dataset into 70\% for training and 30\% was held out for testing. In each dataset, the detectors run on training and test set before adding irrelevant features. The anomaly threshold $t$ is set as the anomaly ratio for each dataset. The detectors performances are demonstrated in Table \ref{tab:realDataSpecs}. 

\subsection{Feature Importance Alternatives}
\label{sec:baselines}

We compare the original PROTEUS system, employing feature selection methods (call it $\text{PROTEUS}_\mathit{fs}$), with the PROTEUS pipeline instantiated only with feature importance methods from related explanation methods. We note that these alternatives have been developed to provide {\em descriptive} explanations; within the PROTEUS pipeline, they are coupled with a classification model, hyper-parameter values are optimized, and they are turned into predictive explanations. 

The research question to study is whether {\em methods specifically developed for explanations in the form of feature importance scores offer additional advantages over the feature selection methods}, everything else being equal (i.e,. the rest of the PROTEUS pipeline). All alternative methods produce {\em local} explanations, i.e., for individual samples. Importance scores for a given feature are calculated for each sample (local scores). We compute the local scores only for the anomalous samples. To incorporate them into PROTEUS and select features for global explanations, the local scores are averaged out for each feature to produce a final feature importance score, as proposed in \cite{Lundberg2020FromLE}. As a final feature selection, we select the top-$K$ features with the highest importance scores. In our experiments, $K$ is set to 10, which is the maximum number of features allowed to be selected by PROTEUS$_\mathit{fs}$ and the feature importance methods. Regarding the hyper-parameters for the feature importance alternatives, we used the ones proposed by the respective authors. We evaluate the following alternatives:

\noindent (1) Lightweight On-line Detector of Anomalies or {\bf LODA}, hereafter, \cite{Pevn2015LodaLO} is an anomaly detector that also returns local feature importance scores. LODA is included as it has shown an excellent trade-off between computational efficiency and anomaly detection performance as a detector \cite{ManzoorLA18}. 
As a feature importance method is selected as a {\bf representative of a detector-specific explanation method}. As such, the results of its explanation method are shown only for the experiments where LODA is also used as the detector. We should stress that when comparing with LODA, the objective is to approximate its performance as the explanation is strongly coupled to the detection process. The resulting PROTEUS variant is called $\text{PROTEUS}_\mathit{LODA}$.

\noindent(2) Kernel {\bf SHAP} (stands for SHapley Additive exPlanations) \cite{LundbergL17} is a model-agnostic method for local explanation of predictive models producing local feature scores. It is considered state-of-the-art, having outperformed LIME \cite{Ribeiro0G16}. As Kernel SHAP does not produce a predictive model itself we consider it as a descriptive method. We use the proposed kernel as in the original publication of SHAP. Kernel SHAP is included as a representative of a {\bf model-agnostic feature importance} method, leading to the variant $\text{PROTEUS}_\mathit{SHAP}$.

\noindent(3) \textbf{CA-Lasso} \cite{MicenkovaNDA13}, is a representative of a {\bf model-agnostic, local feature importance specifically pertaining to anomaly explanation}. It selects $k$-nearest neighbors per outlier $a_i$ and $k$ other random samples. To overcome the class imbalance, the authors oversample $a_i$ adding pseudo-samples around it, labelling them as anomalies by assumption, until the two classes are balanced. The explanation problem is then turned into binary classification per outlier solved with Lasso. The feature importance of each feature for $a_i$ corresponds to the Lasso coefficients. 
Rather than learning the decision boundary of individual anomalies PROTEUS builds a binary classifier to explain all the anomalies spotted by an unsupervised detector. In that sense, feature selection in \cite{MicenkovaNDA13} generates local explanations per anomaly that do not generalize to unseen anomalies. Moreover, PROTEUS oversampling is supervised (by the detector) while numerous feature selection and classification algorithms along with the optimization of their hyper-parameter values.  Finally, out-of-sample (predictive) performance is estimated by PROTEUS using AUC for subset selection instead of accuracy as originally proposed in \cite{MicenkovaNDA13}. The resulting PROTEUS variant is called $\text{PROTEUS}_\mathit{CA-Lasso}$.

\subsection{PROTEUS Performance Estimation}
\begin{figure}[t]
    \centering
    \includegraphics[width=1\linewidth]{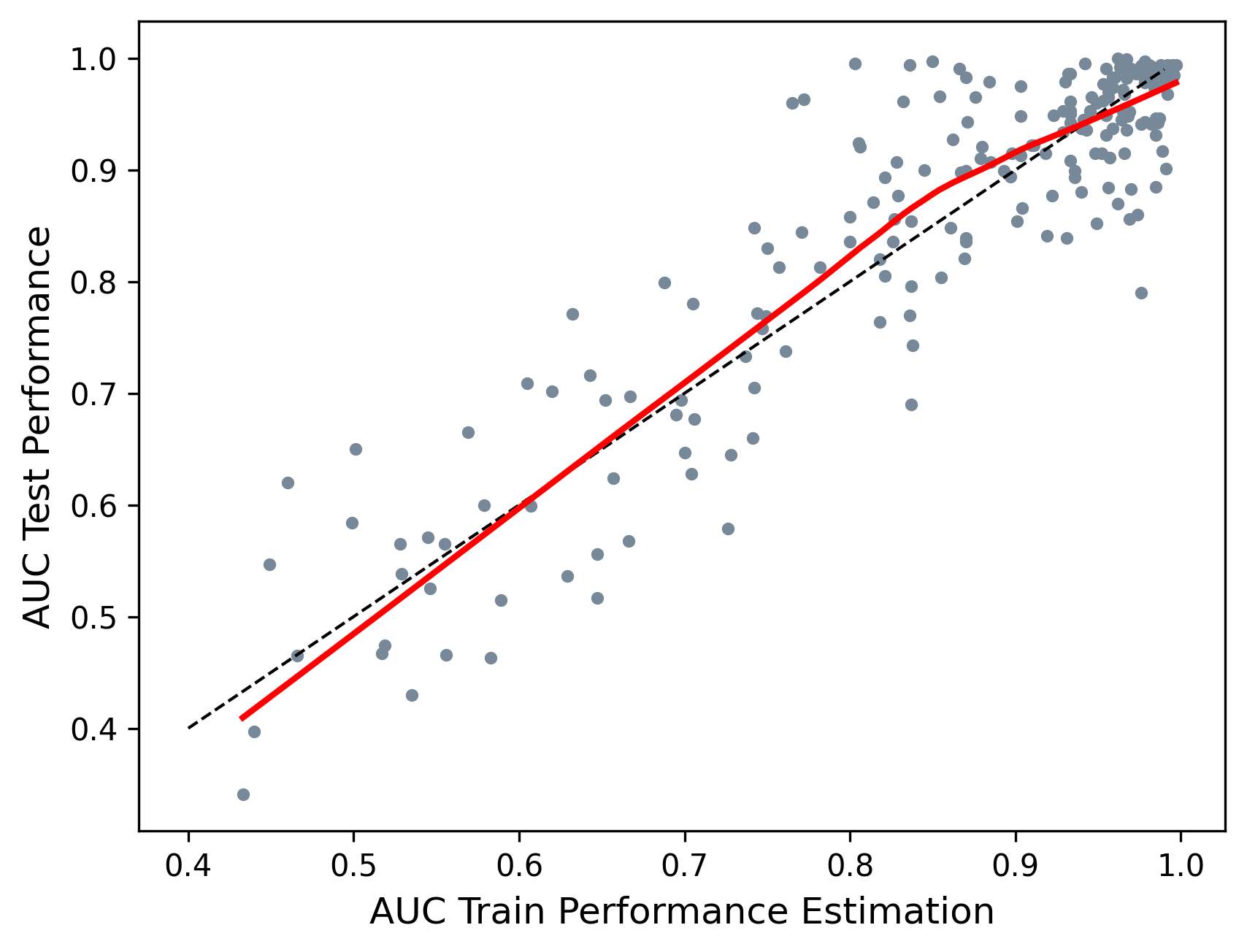}
    \caption{Bias between train and test AUC performances of PROTEUS implemented with BBC and CV with grouping. 
    }
    \label{fig:biasFigProt}
\end{figure}

The objective of this experiment is to assess the effect of PROTEUS design choices, specifically the BBC and Grouping, to provide an accurate performance estimation. Figure \ref{fig:biasFigProt} depicts the train estimates and test performance when PROTEUS is employed with the design choices described in Section \ref{sec:proteusPipeline}, i.e., BBC and CV with Grouping. The dashed black diagonal line indicates the zero bias: points above the diagonal indicate underestimation (negative bias) and below overestimation (optimistic bias). To show the accuracy of the estimation of PROTEUS design choices, we fit a loess curve\footnote{https://en.wikipedia.org/wiki/Local\_regression} on train and test performances for every combination (258 in total) of datasets (synthetic and real), detectors (IF, LOF and LODA) and feature selection methods (general purpose and feature importance methods). Ideally, we would want the loess curve to fit exactly the diagonal. Observe that with lower AUC performances PROTEUS tends to overestimate while with higher performances PROTEUS returns a more conservative estimation. In both cases, the points are close to the ideal diagonal line. 

\begin{figure}[t]
    \centering
    \includegraphics[width=1\linewidth]{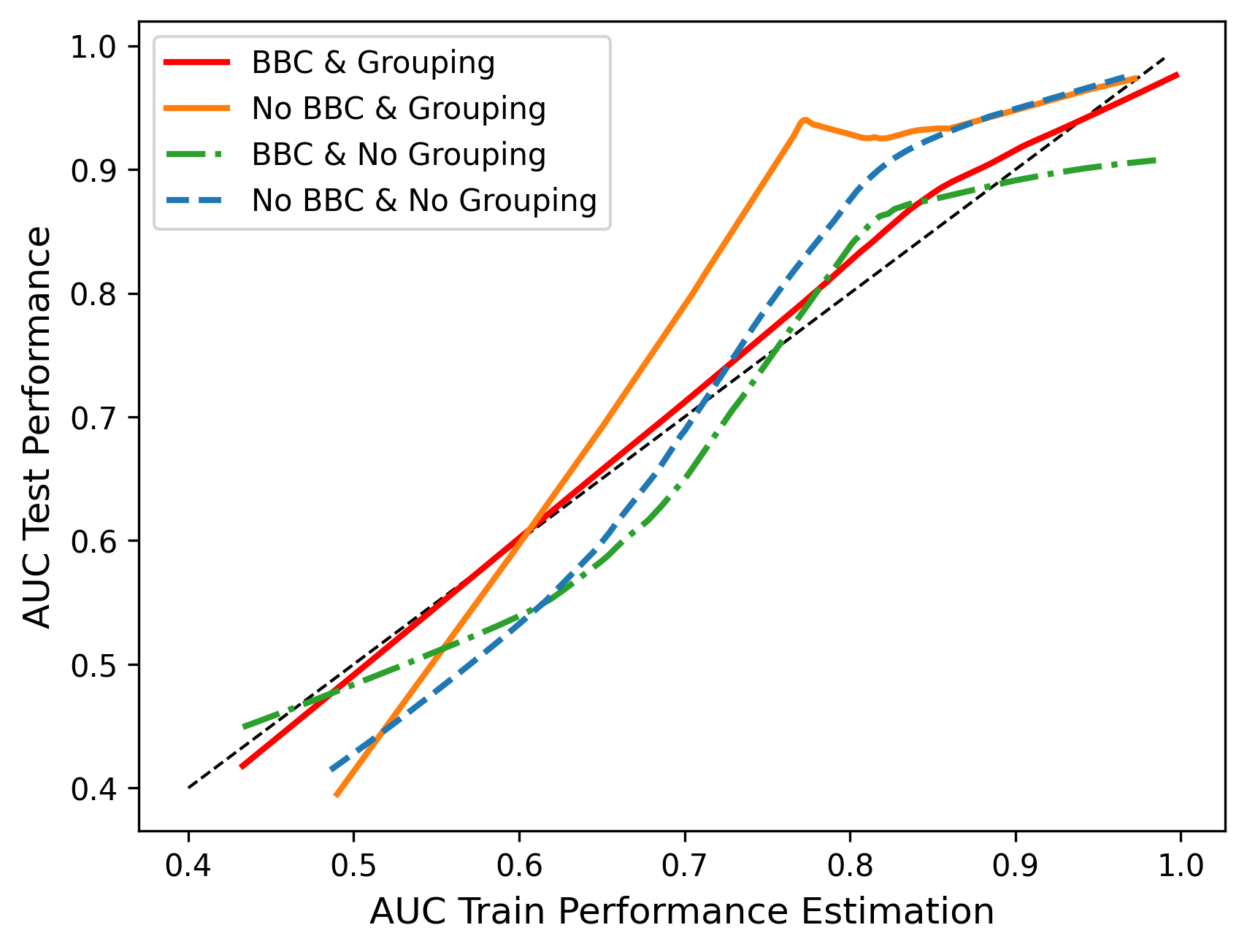}
    \caption{Bias between train and test AUC performance of PROTEUS implemented with 4 alternatives.}
    \label{fig:biasFig}
\end{figure}

To further show the efficacy of the proposed design choices to provide an accurate performance estimation, in Figure \ref{fig:biasFig} we compare the loess curves for train and test estimates for (i) BBC and Grouping (our design choices), (ii) no BBC (i.e., CV estimate) and Grouping (iii) BBC and no Grouping and (iv) no BBC and no Grouping. To quantify the bias for each of the four alternatives, we use the Residual Sum of Squares (RSS) to measure the discrepancy between the train and test performance. When PROTEUS is employed with BBC and Grouping (i), it gives the most accurate estimation of out-of-sample performance (with $\mathit{RSS_{(i)}}$ = 0.05) than when using any of the three alternative design choices (with $\mathit{RSS_{(ii)}}$ = 0.88, $\mathit{RSS_{(iii)}}$ = 0.11 and $\mathit{RSS_{(iv)}}$ = 0.25). 

\begin{figure*}
    \centering
    \includegraphics[width=0.7\linewidth]{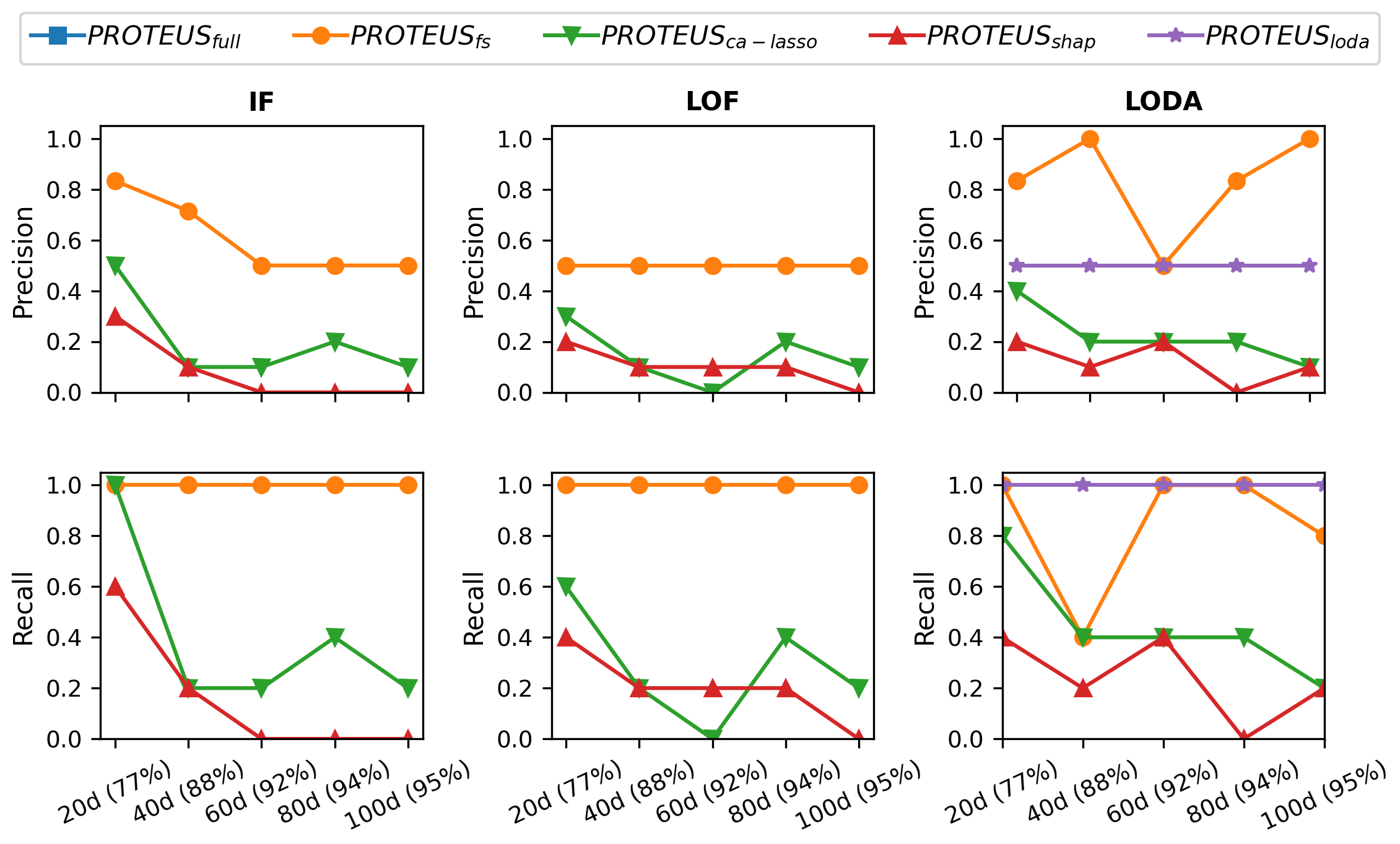}
    \caption{Precision and Recall performance of  discovered features when explaining IF, LOF and LODA on synthetic datasets w.r.t. increasing data dimensionality (\% irrelevant feature ratio)}
    \label{fig:synthPrecRec}
\end{figure*}

\begin{figure}
    \centering
    \includegraphics[width=1\linewidth]{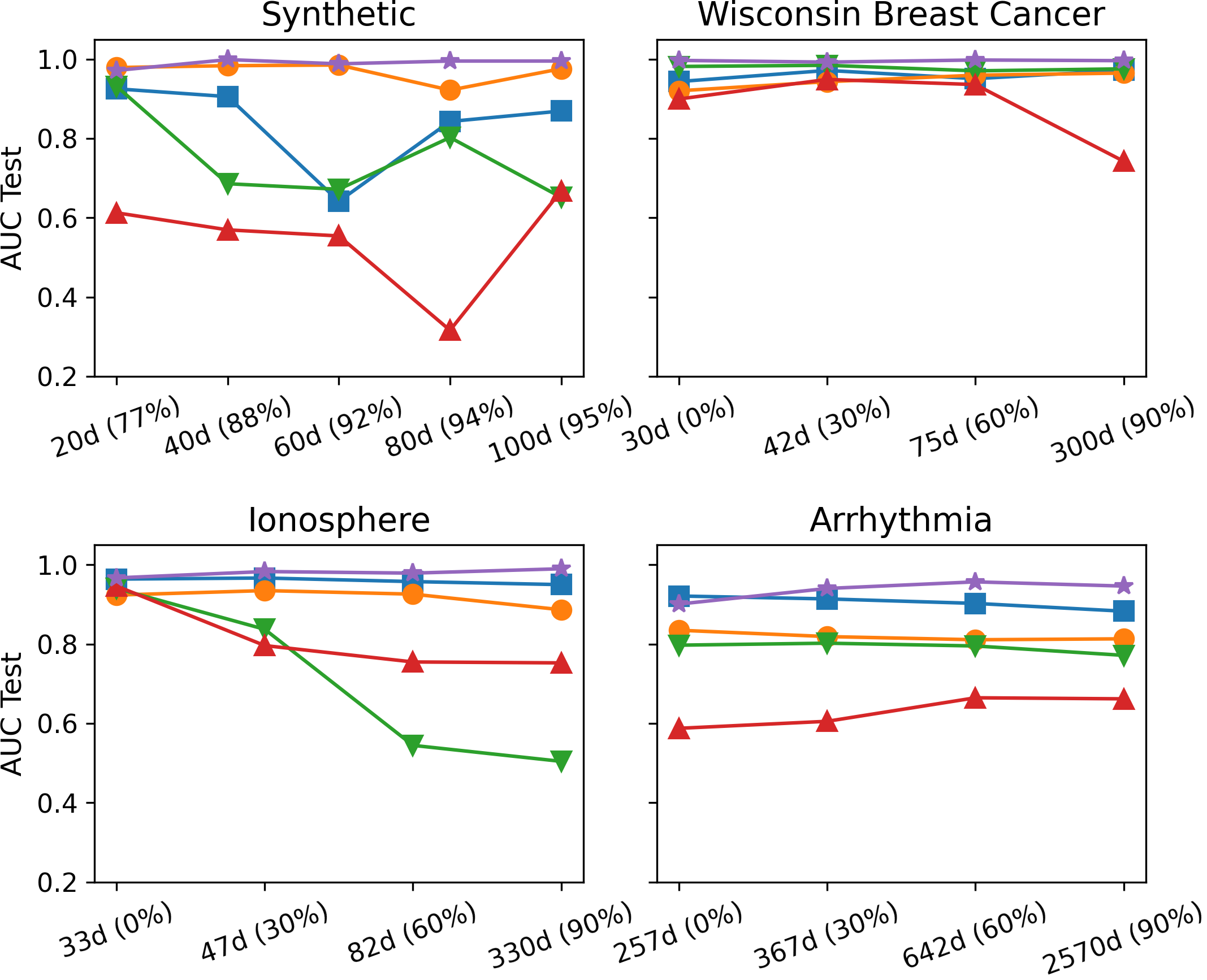}
        \caption{AUC test performance averaged over the three detectors on synthetic and real datasets w.r.t. increasing dimensionality (\% irrelevant feature ratio)}
        \label{fig:dimAUC}
\end{figure}

\subsection{Relevant Features Identification Accuracy}
The goal of this experiment is to verify whether the features discovered during the training phase by PROTEUS$_\mathit{fs}$ and the feature importance alternatives are part of the gold-standard feature subset $S$. For this experiment we used the \emph{synthetic} datasets. To assess the quality of the global explanation $E$ in terms of features, we compute $precision(S, E) = \frac{|S ~ \cap ~ E|}{|E|}$ and $recall(S, E) = \frac{|S ~ \cap ~ E|}{|S|}$. As we select the top-10 features to form the explanation and $S$ contains 5 features, the $precision$ for the feature importance alternative methods will be up to 0.5. The recall and precision curves are depicted in Figure \ref{fig:synthPrecRec}. Feature selection methods employed by PROTEUS$_\mathit{fs}$ exhibit the highest precision never dropping below 0.5, independently of the employed detector or dataset dimensionality. We observed that precision is 0.5 when Lasso is selected and higher when FBED is selected. We should stress that SES was never selected by PROTEUS for the synthetic datasets. FBED removed most of the irrelevant features leading to a predictive model with less than 10 features to approximate the decision boundary of the corresponding detector. PROTEUS$_\mathit{fs}$ achieves almost optimal recall regardless of the dimensionality and the employed detector. A slight drop in recall is observed when the precision higher than 0.8 (achieved only by FBED), while recall is optimal when Lasso is selected. Moreover, PROTEUS$_\mathit{fs}$ feature selection methods are robust to increasing data dimensionality and irrelevant feature ratio where CA-Lasso and SHAP seem to be particularly sensitive. 


\subsection{PROTEUS Generalization Performance}
The objective of this experiment is to assess the generalization performance of PROTEUS without (PROTEUS$_\mathit{full}$) and with feature selection (PROTEUS$_\mathit{fs}$) as well as with the various feature importance alternatives, ($\text{PROTEUS}_\mathit{CA-Lasso}$,  $\text{PROTEUS}_\mathit{SHAP}$, $\text{PROTEUS}_\mathit{LODA}$). 
Figure \ref{fig:dimAUC} depicts the AUC performance for each method in test set. Regarding the synthetic datasets, PROTEUS$_\mathit{fs}$ achieves very high AUC across the increasing data dimensionality with a minimum of 0.96. CA-Lasso and SHAP instead exhibit lower performances as they do not retrieve, as showed in the previous experiment, many of the relevant features. Observe that in the synthetic dataset PROTEUS$_\mathit{fs}$ generalizes better than PROTEUS$_\mathit{full}$, i.e., when using all the available features. 

Regarding the real datasets, similar trends are observed with PROTEUS$_\mathit{fs}$ achieving consistently a very high generalization performance with a minimum of 0.8 in Arrhythmia in the presence of 2,570 dimensions and 90\% irrelevant feature ratio. PROTEUS$_\mathit{fs}$ seems to approximate in a detector-agnostic manner, the optimal performance of LODA's feature importance method when LODA is used as the detection algorithm. This is due to the fact that LODA's explanations are tailored to its detection algorithm; however, if LODA's detection performance was poor in a dataset, the provided explanation would be of less value for the analysts. The feature selection methods employed by PROTEUS$_\mathit{fs}$, are able to discover the relevant features leading to predictive models with very high performance regardless of the data dimensionality (and the increasing relevant feature ratio) and capture accurately the decision boundary of every employed unsupervised detector. 


\subsection{PROTEUS RunTime Performance}

\begin{figure}[t]
    \centering
    \includegraphics[width=.9\linewidth]{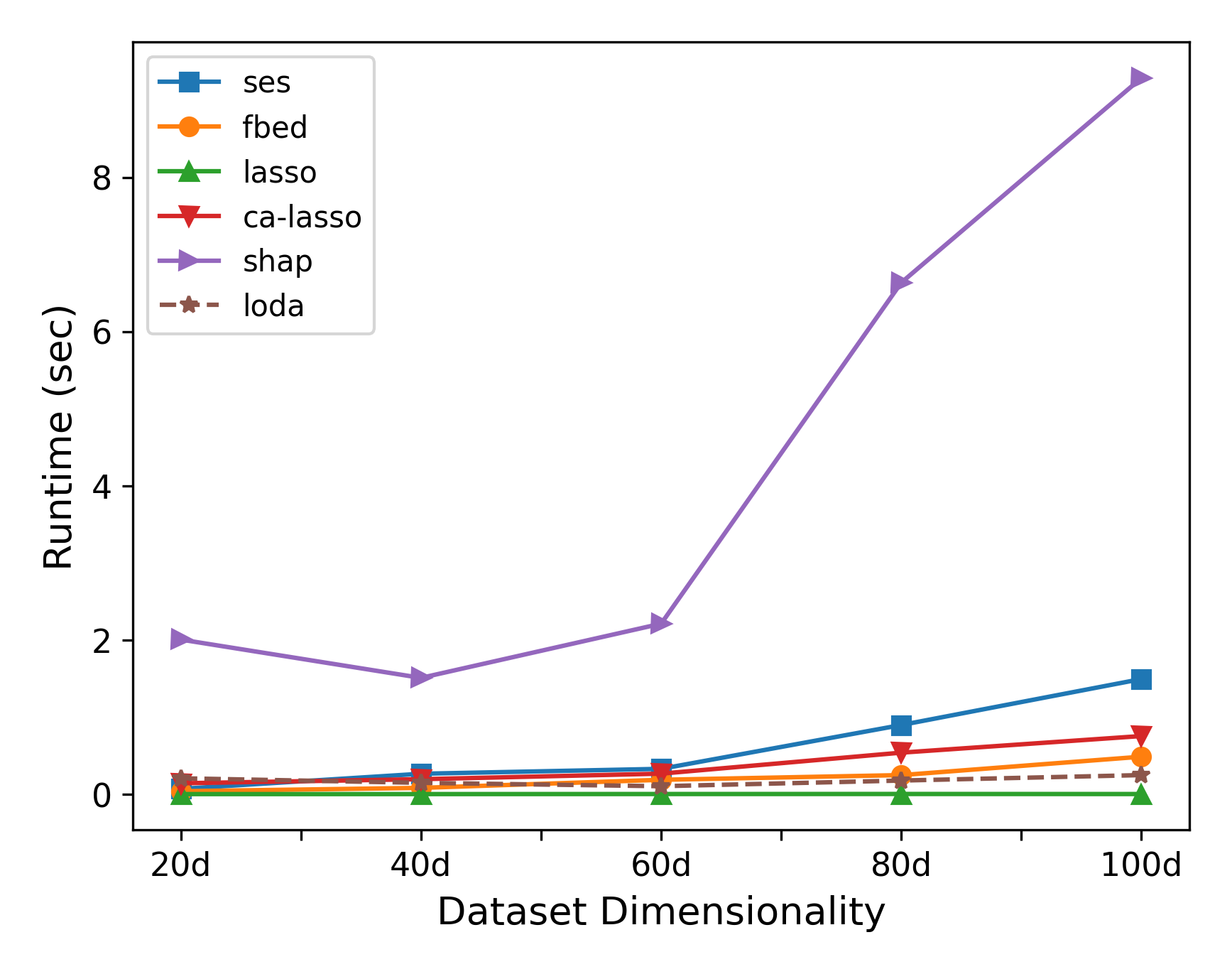}
    \caption{Average runtime of feature selection/importance methods on synthetic datasets of increasing dimensionality.}
    \label{fig:runtimeFselMethods}
\end{figure}

In the subsequent experiment, we demonstrate the execution time of the feature selection methods employed by PROTEUS. Figure \ref{fig:runtimeFselMethods} depicts the runtime comparison between the ad-hoc feature importance methods and the feature selection algorithms. The employed methods are specifically designed to search efficiently the exponential search space and thus require less than two seconds on average in 100-dimensions to select features, exhibiting a steady execution time. On the contrary, SHAP is  the most expensive method; as we had to explain the outcome of any employed detector, we used Kernel SHAP which is model-agnostic. Given the fact that SHAP is optimized only for particular families of algorithms, e.g. tree-based, its execution time is particularly sensitive to data dimensionality because the Shapley values must be calculated for all the input features. Recall that in PROTEUS we tried three classifiers resulting 30 combinations according to their hyper-parameters and three feature selection algorithms resulting 20 combinations according to their hyper-parameters including the full selector, i.e., when the full feature space is considered. Thus, the total number of configurations tried in PROTEUS is 600. Each configuration requires 2 seconds on average to complete regardless of the dataset dimensionality.

\section{Contrasting PROTEUS Surrogate Models with Unsupervised Anomaly Detectors}
\label{sec:qualityAssessment}

In this section, we are investigating possible disagreements between PROTEUS's surrogate models and unsupervised Anomaly Detectors, namely detected anomalies explained as normal points or vice-versa.

To assist human analysts in spotting such suspicious samples, we introduce an original visualization method based on \emph{spider charts}. The proposed \emph{Spider Anomaly Explanation (SAE) charts} is essentially a 2D visualisation of multivariate data projected over the explaining subspaces returned by PROTEUS. 
The chart has a ``web-like'' form with concentric circles and several spokes where each one corresponds to a specific feature. Extreme values of the features are depicted near the center or near the outermost circle. Then, a mutli-dimensional sample is represented as an irregular polygon intersecting every spoke according to the quantile its feature values falls in. 

In this work, we propose a variation of spider charts tailored to anomaly explanation. First, instead of plain feature values, we consider each concentric circle in the chart to represent one of the four quantiles where the center corresponds to the 0 quantile and the outermost circle corresponds to the 1 quantile. Then, every feature value is translated to a quantile ranging from 0 to 1. Hence, the normal region in the chart is the interquartile range (IQR) containing 50\% of the values. Finally, we reverse the samples with extreme low values belonging to quantiles 0 - 0.25 to the 0.75 - 1 quantiles so that both low and high extremes can be identified near the outermost circle, far from the normal region. When a sample's value intersect with a spoke in the quantiles 0.75 - 1, it means that at least 75\% of values for the particular feature fall below the sample's value.

\begin{figure}[h]
    \centering
    \includegraphics[width=1.0\linewidth]{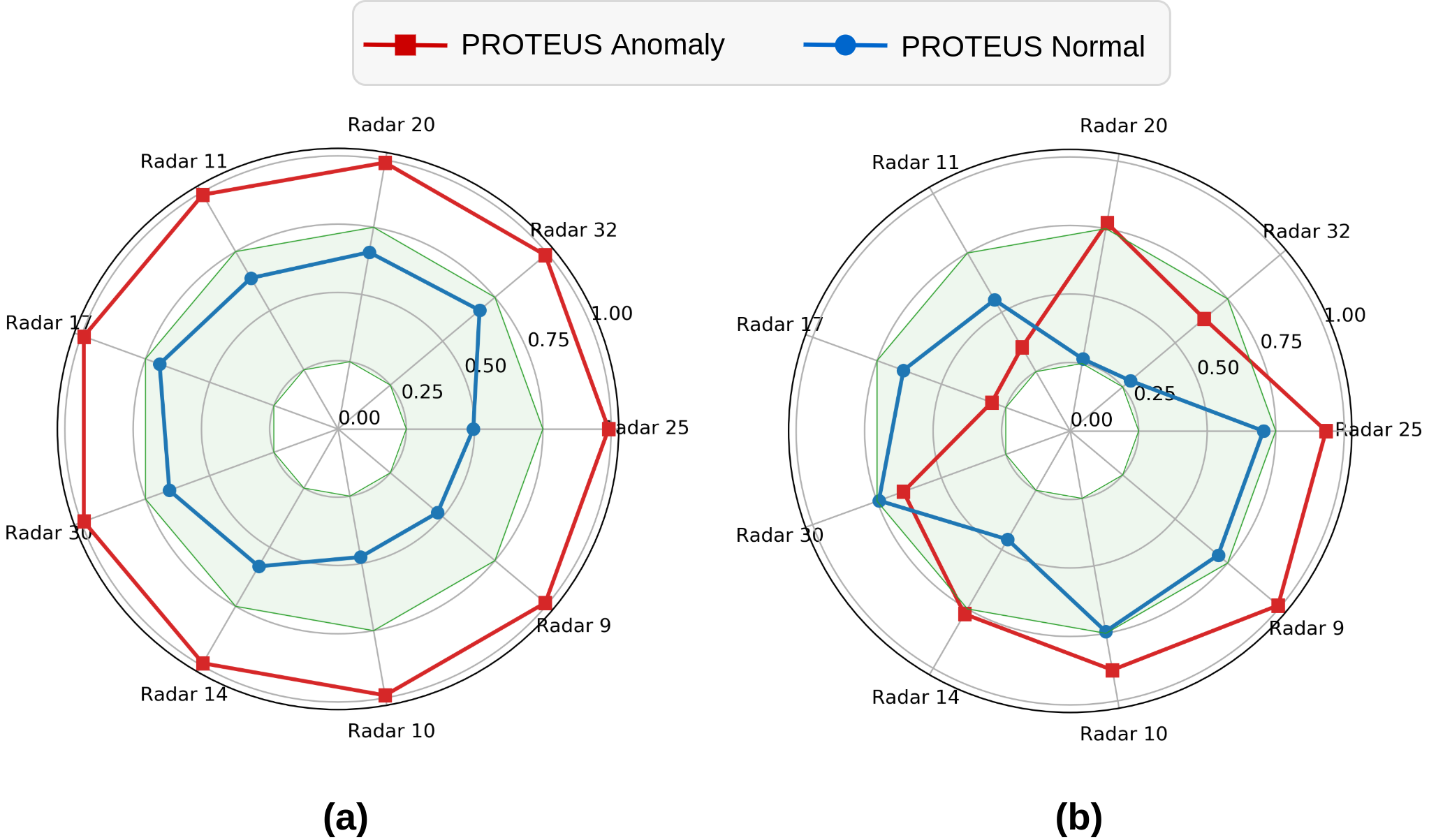}
    \caption{Spider Anomaly Explanation Charts when explaining LOF in Ionosphere using PROTEUS. 
    }
    \label{fig:spiderCharts}
\end{figure}

In Figure \ref{fig:spiderCharts} we demonstrate two SAE charts when explaining the LOF detector in the Ionosphere dataset. The explanation produced by PROTEUS comprises of 9 features. In Figure \ref{fig:spiderCharts}a both PROTEUS's surrogate model and LOF agreed on the labels of these two samples. We can observe that the normal sample falls entirely into the normal (green) region while the anomalous sample deviates significantly in every feature. Figure \ref{fig:spiderCharts}b illustrates two samples where PROTEUS's surrogate model disagrees with LOF on their labels. LOF  identified the blue sample as anomaly while PROTEUS identified it as normal. We can clearly observe that this sample was erroneously detected by LOF as it falls entirely into the normal region. For the other conflict (the red sample) it is not as obvious as in the former case because it deviates w.r.t. a subset of the features of the explanation. This sample is an anomaly according to the gold standard that was not detected by LOF. However, PROTEUS considered this sample an anomaly, extracting three features (Radar 10, 9 and 25) that it takes extreme values. We should finally stress that since PROTEUS strives to explain all the anomalies simultaneously, it is likely that an anomalous sample deviates w.r.t. a subset of the explaining subspace.

To quantify the utility of a PROTEUS explanation to reveal errors made by an unsupervised detector we introduce two metrics that rely on the gold standard available for each dataset. We consider as \emph{conflicts} the suspicious samples for which the PROTEUS's surrogate model predicts a different label than the detector. Subsequently, we define two sets of conflicts following the notation of Section \ref{sec:problemdef} where $\omega'_A$ is the detector model, $f(\cdot, \theta^*)$ is the PROTEUS's surrogate model equipped with the best found hyper-parameters, ``1'' denotes an anomaly and ``0'' denotes a normal sample.

\begin{definition}
\emph{Anomaly Normal Conflicts (ANC)}: Each sample that the detector model labels as anomaly while PROTEUS's surrogate model labels as normal.
\[
\mathit{ANC} = \{ s ~|~ \omega'_A(s) = 1 \wedge f(s, \theta^*) = 0\}, 
\]
\end{definition}

\begin{definition}
\emph{Normal Anomaly Conflicts (NAC)}: Each sample that the detector model labels as normal while PROTEUS's surrogate model labels as anomaly.
\[
\mathit{NAC} = \{ s ~|~ \omega'_A(s) = 0 \wedge f(s, \theta^*) = 1\}
\]
\end{definition}

\noindent
Based on the two previous sets we define two metrics to quantify the utility of a PROTEUS explanation.

\begin{definition}
\emph{True Normal Discovery (TND)}: The ratio of conflicted samples that PROTEUS's surrogate model labelled correctly as normals according to the $\mathit{True~Normals}$ in the gold standard.
\[
\mathit{TND} = \frac{|\mathit{ANC} \cap \mathit{True~Normals}|}{|\mathit{ANC}|}
\]
\end{definition}

\begin{definition}
\emph{True Anomaly Discovery (TAD)}: The ratio of conflicted samples that PROTEUS's surrogate model labelled correctly as anomalies according to the $\mathit{True~Anomalies}$ in the gold standard.
\[
\mathit{TAD} = \frac{|\mathit{NAC} \cap \mathit{True~Anomalies}|}{|\mathit{NAC}|}
\]
\end{definition}

When there are no conflicts, i.e., PROTEUS approximates perfectly the detector's decision boundary (AUC = 1 on test set), the two metrics are not defined ($\mathit{ANC}$ and $\mathit{NAC}$ are empty). In case of conflicts, $\mathit{TND}$ and $\mathit{TAD}$ range between 0 and 1. Values close to 1 indicate that PROTEUS' surrogate model disagrees with the detector model and it labels suspicious samples correctly w.r.t. the gold standard. On the contrary, values close to 0 indicate that PROTEUS disagrees incorrectly with the detector. Clearly, the number of conflicts is higher when PROTEUS exhibits low AUC performance. 

\begin{figure}[ht]
    \centering
    \includegraphics[width=.9\linewidth]{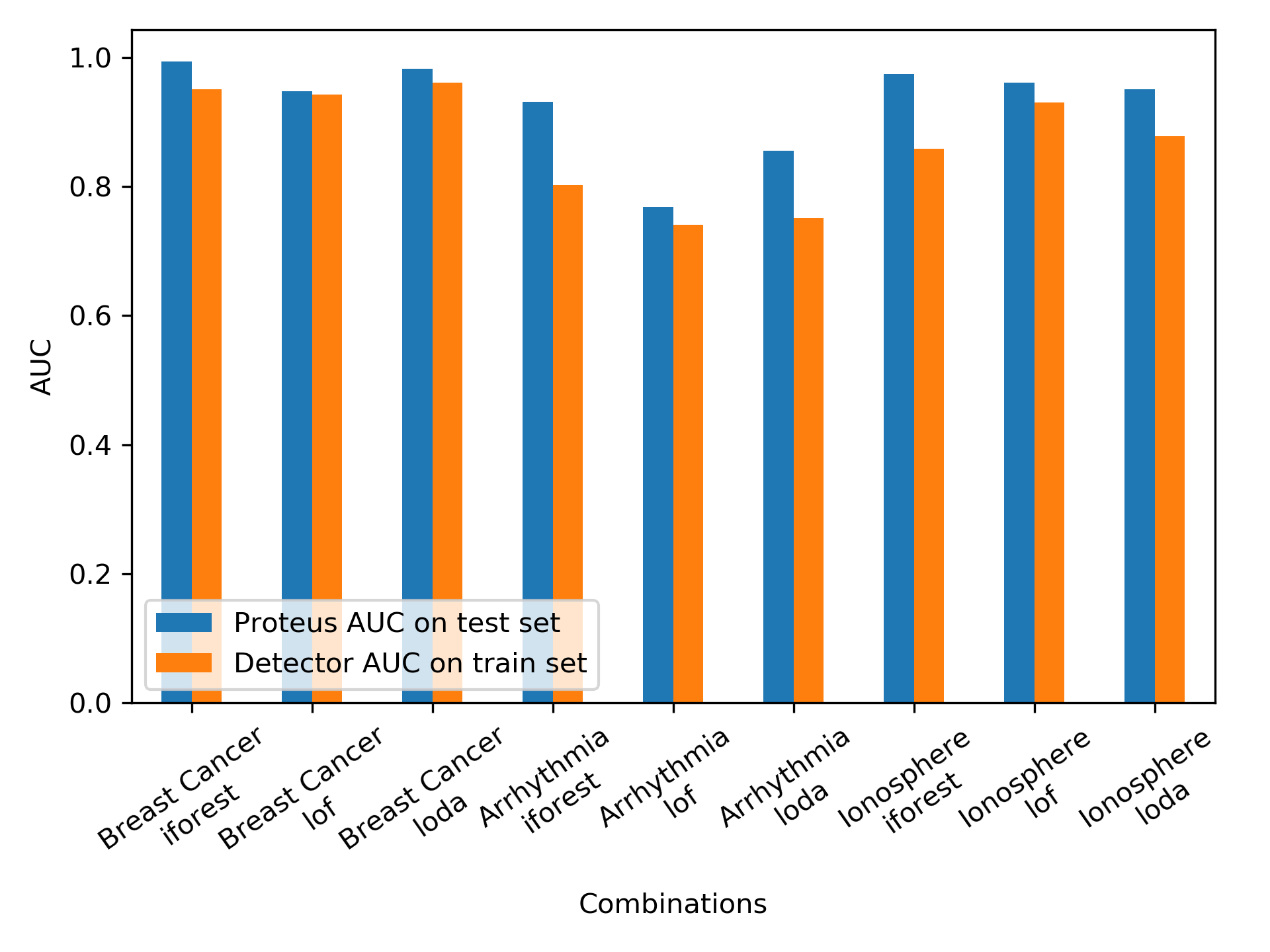}
    \caption{PROTEUS AUC against anomaly detector AUC for real datasets.}
    \label{fig:qaAucComp}
\end{figure}

Figure \ref{fig:qaAucComp} contrasts the AUC of PROTEUS against the AUC of the three detectors used to analyse each real dataset. The former is computed on the test (holdout) set using the labels produced by a detector and serves as the approximation quality of its decision boundary. The latter is computed on the train set using the labels of the gold standard and reveals the effectiveness of a detector to identify anomalies in a dataset. We can easily observe that \emph{the quality of the approximation of a detector's decision surface by PROTEUS decreases as the detector's effectiveness decreases}. For instance, in Arrhythmia we observe the lowest AUCs for the three detectors and also for the surrogate models of PROTEUS. This trend can be attributed to the fact that some misdetected samples are very hard to classify correctly without a very complex boundary. However, if the surrogate model needs to learn a more complex decision surface to segregate the misdetected samples from their neighbors, it makes the surrogate model prone to overfitting and thus reducing its generalization performance. Overfitting is actually avoided thanks to the CV protocol; PROTEUS will strive to select models that generalize well in unseen data optimizing the out-of-sample AUC performance. 

\begin{figure}[ht]
    \centering
    \includegraphics[width=.9\linewidth]{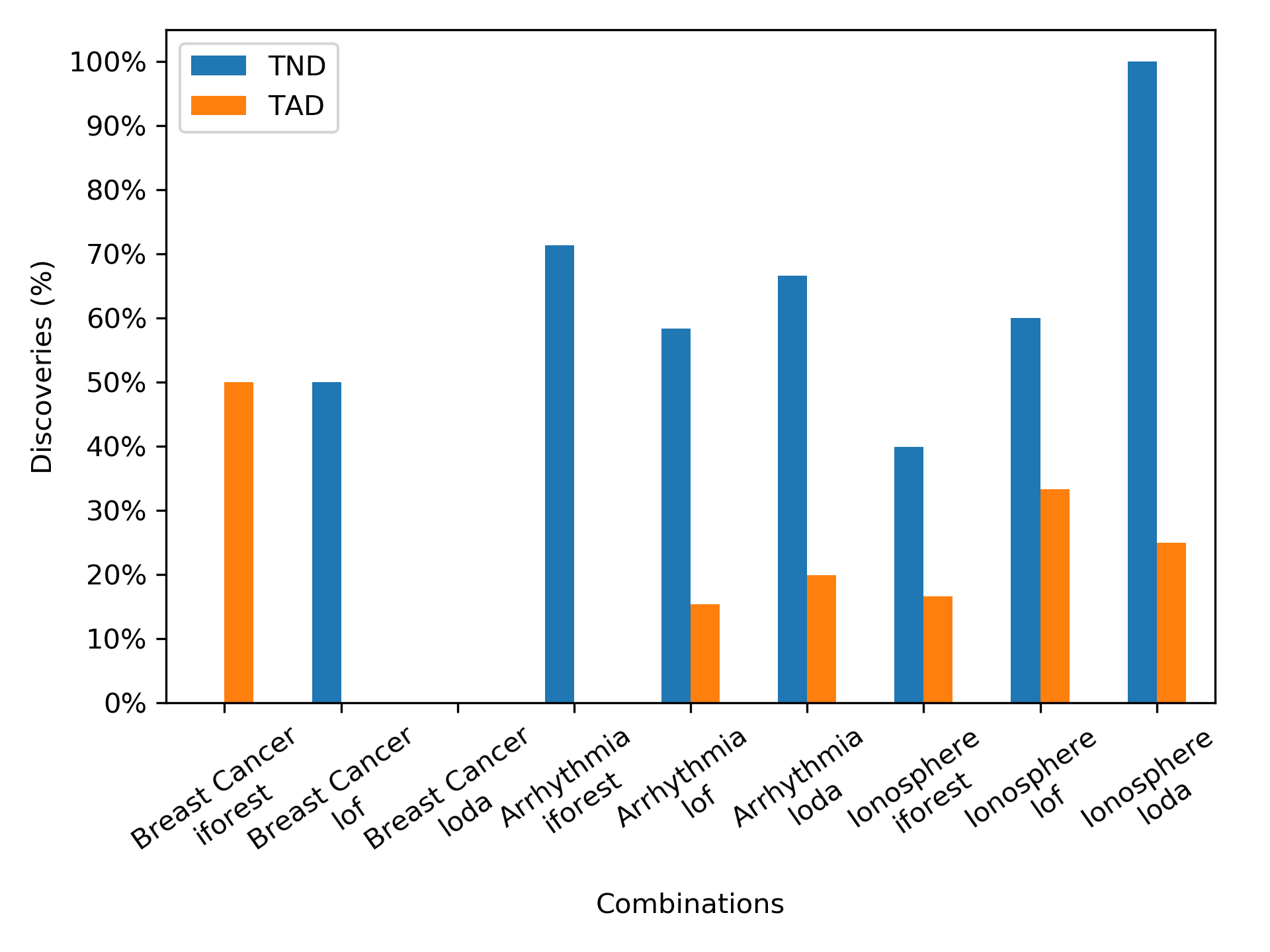}
    \caption{Fraction of samples (y-axis) identified as TND and TAD according to PROTEUS explanations for different combinations of detectors and datasets  (x-axis).}
    \label{fig:qaDiscoveries}
\end{figure}

\begin{figure}[t]
    \centering
    \begin{subfigure}[b]{0.49\linewidth}
        \centering
        \includegraphics[width=\textwidth]{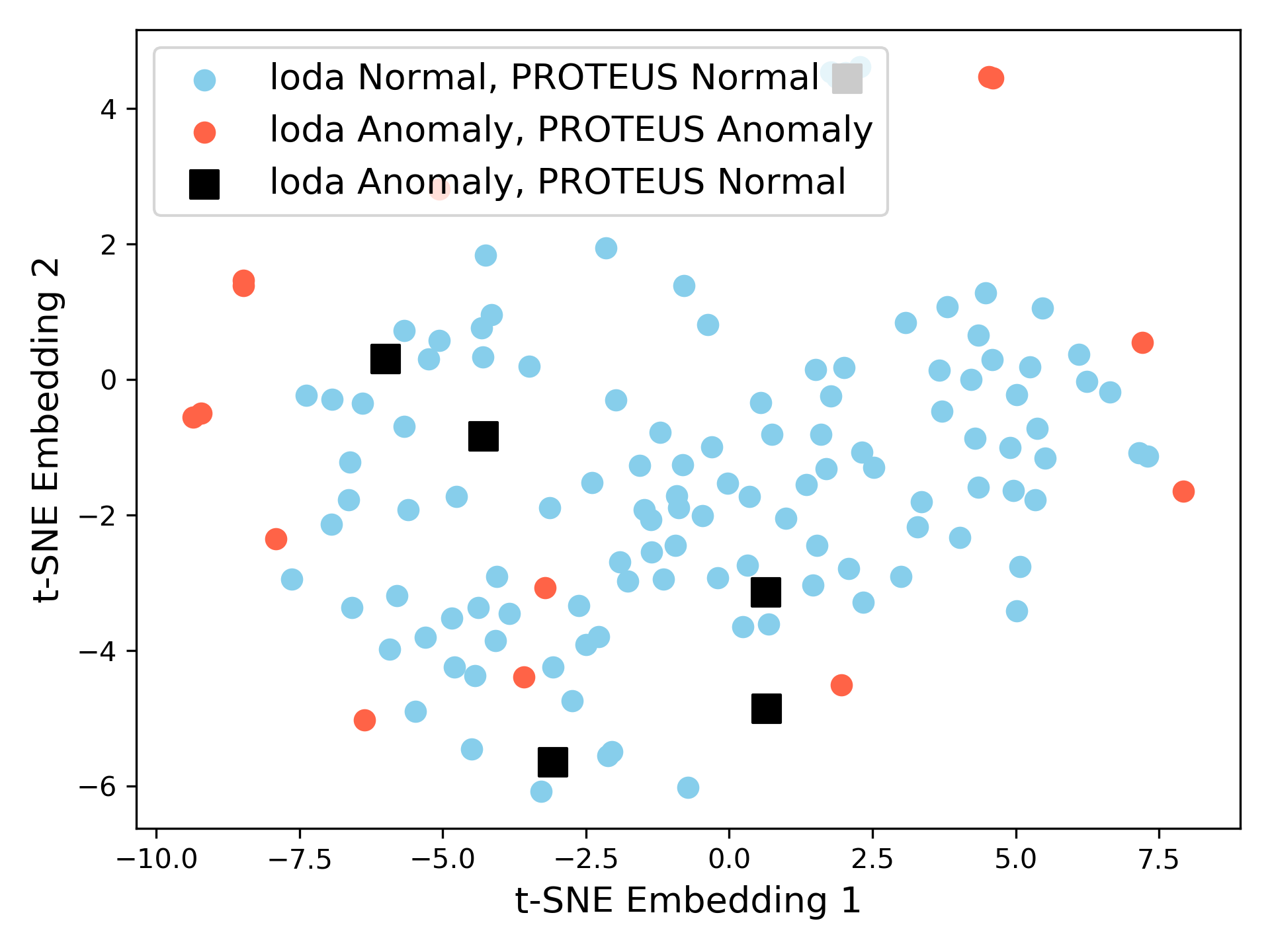}
        \caption{}
        \label{fig:tsneANC}
    \end{subfigure}%
    \hfill
    \begin{subfigure}[b]{0.49\linewidth}
        \includegraphics[width=\textwidth]{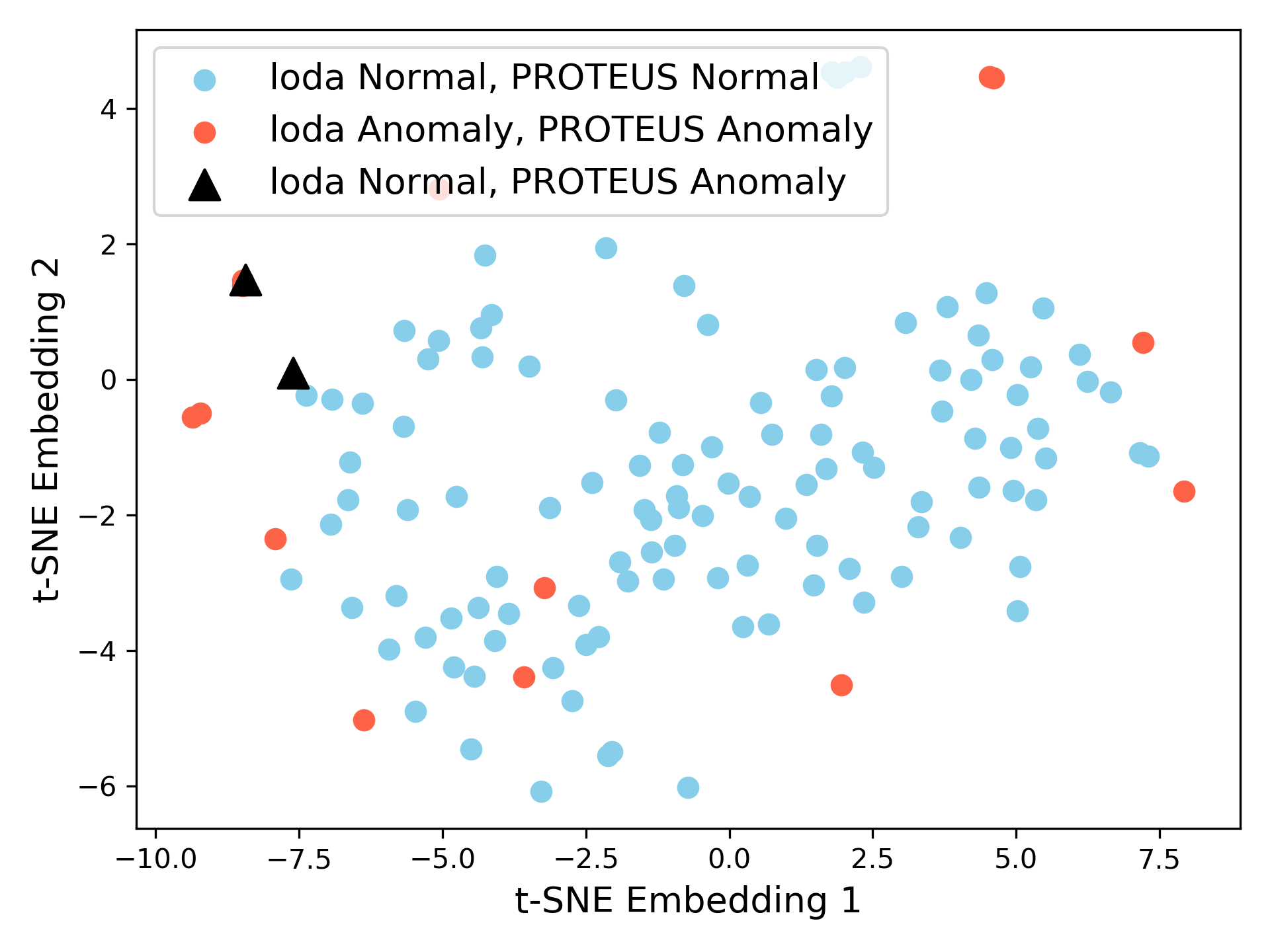}
        \caption{}
        \label{fig:tsneNAC}
    \end{subfigure}
    \caption{A 2-D reduced t-SNE visualisation of Arrhythmia according to PROTEUS 10-D explanation for LODA. 
    }
    \label{fig:tsneVis}
\end{figure}

Figure \ref{fig:qaDiscoveries} sheds some light on the percentage of conflicting samples between PROTEUS and unsupervised detectors per dataset. PROTEUS reveals more True Normals with an average TND $\sim 50\%$ than True Anomalies with an average TAD $\sim 18\%$. In other words, \emph{PROTEUS seems to be more effective in discovering false alarms}. To justify this claim we consider a 2D reduced visualization using t-SNE \cite{Maaten2008VisualizingDU} of the Arrhythmia dataset projected over 10-dimensional PROTEUS explanation for LODA. Figures \ref{fig:tsneNAC} and \ref{fig:tsneANC} depict the agreements between PROTEUS and LODA as circles and their disagreements as rectangles for \emph{ANC} and triangles for \emph{NAC}. Figure \ref{fig:tsneNAC} illustrates the \emph{ANC} samples contributing to the identification of \emph{True Normals}. 
As expected, these samples are located within dense regions, surrounded by normal samples, requiring more complex boundaries to separate. On the contrary, Figure \ref{fig:tsneANC} illustrates the \emph{NAC} samples contributing to the identification of \emph{True Anomalies}. These samples lie on sparse areas where less complex boundaries can be built to separate them. This is because less complex boundaries enable better generalizing models and thus PROTEUS can classify misdetected normals in sparse areas, not yielding many \emph{True Anomalies}.

To conclude PROTEUS constructs a reduced-dimensionality surrogate model that not only generalizes well to unseen samples but also provides valuable insights for identifying False Negatives and False Positives of unsupervised anomaly detectors. 

\section{Related Work}
\label{sec:relWork}

In this section, we survey various categories of related work on explaining anomalies in unsupervised and supervised settings, partially inspired by \cite{MachaA18}. We should stress that explanations of anomalies in temporal data is beyond the scope of this work \cite{Giurgiu2019,Bessa2020}.

\subsection{Explainable Anomaly Detectors}

As unsupervised detectors assess the abnormality of multidimensional data on various feature subspaces, they can also report the subspaces that contributed the most to the anomaly score of a particular sample. A first example of such explainable anomaly detectors is LODA \cite{Pevn2015LodaLO} which scores samples based on the average log density over an ensemble of one-dimensional histogram density estimators. Given that each histogram (with sparse projections) scores a randomly generated subspace, LODA explanations are essentially a list of features ranked according to their contribution to the anomalousness score of a sample.

LODI \cite{DangMAN13} and LOGP \cite{DangANZS14} seek an optimal subspace in which an anomaly is maximally separated from its neighbors. Both works exploit a dimensionality reduction technique to measure the anomalousness of a sample in a low-dimensional subspace capable of preserving the locality around its neighbors while at the same time maximizing its distance from this neighborhood. Then, the explanation of a sample is the top-k features with the largest absolute coefficient from the eigenvector with the largest eigenvalue.

\cite{SiddiquiFDW19} proposes an interactive explanation method that can be used for any density-based anomaly detector.  \cite{KriegelKSZ09} introduced a method to detect anomalies in axis-parallel subspaces, called SOD, that computes the anomaly score of a  in a hyperplane w.r.t. to nearest neighbors in the full space. SOD hyperplanes that contribute most in the anomaly scores serve as explanations. CMI \cite{BohmKMNV13} and HiCS \cite{KellerMB12} rely on statistical methods to select subspaces of high-dimensional datasets, where anomalies exhibit a high deviation from normal samples. Both consider highly contrasting subspaces as explanations of all possible anomalies in a dataset. 

\textit{The previous works explain anomalies as a byproduct of an unsupervised detection method. Given that independent experimental evaluations showed that no detector outperforms all others for all possible datasets \cite{Goldstein2016,Campos2016,Domingues2018,Xiaoyi2019}, in our work we focus on learning the decision boundary of any unsupervised anomaly detector that could be used for a particular dataset. In contrast to the descriptive explanations provided by the aforementioned works, PROTEUS targets predictive explanations that could be successfully used to detect and explain anomalies also in unseen data.}

\subsection{Post-hoc Anomaly Explainers}

The primary focus of these methods is to specify a subset of features such that a sample may obtain a high anomaly score when projected onto these subspaces. Some authors have referred to this explanation task as “outlying aspects mining” \cite{Duan2014MiningOA,VinhCBLRP15}. 

We first consider works providing local explanations. The seminal work \cite{KnorrN99} first introduced the problem of explaining individual outliers with “Intensional knowledge" under the form of minimal feature subspaces in which they show the greatest deviation from inliers. To find optimal subspaces, \cite{KuoD16} formulates a constraint programming problem that maximizes differences between neighborhood densities of known outliers and inliers. \cite{KellerMWB13} employs a search strategy aiming to find a subspace which maximizes differences in anomaly score distributions of all samples across subspaces while \cite{MicenkovaNDA13} measures the separability per anomaly using classification accuracy, and then apply Lasso to produce a local explaining subspace. OAMiner \cite{Duan2014MiningOA} finds the most outlying subspace where a sample is ranked highest in terms of a probability density measure and OARank \cite{VinhCBLRP15} ranks features based on their potential contribution toward the anomalousness of a sample. 
\textit{Rather than learning the decision boundary of individual anomalies, PROTEUS builds a classifier to explain simultaneously all the anomalies spotted by an unsupervised detector. Moreover, PROTEUS's oversampling is supervised, optimizing the hyper-parameters of various feature selection and classification algorithms.}  

Extending earlier work \cite{AngiulliFP09} on explaining individual anomalies, \cite{AngiulliFP13,AngiulliFPM13} focus on explaining groups of anomalies for categorical data using contextual rule based explanations. Authors search for $<$context, feature$>$ pairs, where the (single) feature can differentiate as many outliers as possible from inliers while sharing the same context. The anomalousness score of a sample in a subspace is calculated based on the frequency of the value that the outlier takes in the subspace. It tries to find subspaces E and S such that the outlier is frequent in one and much less frequent than expected in the other. To avoid searching exhaustively all such rules, the method takes two parameters, and, to constrain the frequencies of the given sample in subspaces E and S, respectively. Similarly, \cite{ZhangDM17} describes anomalies grouped in time. They construct explanatory Conjunctive Normal Form rules using features with low segmentation entropy, which quantifies how intermixed normal and anomalous samples are. They heuristically discard highly correlated features from the rules to get minimal explanations. The previous related work assumes that anomalies are scattered and strive to explain them individually rather than to summarize the explanation of a collection of anomalies. 

The following works perform explanation summarization aiming to explain a set of anomalies collectively rather than individually. LookOut \cite{Gupta2018BeyondOD} exploits a submodular optimization function to ensure concise summarization. xPACS \cite{MachaA18} groups anomalies by generating sequential feature-based explanations providing a ranked list of feature-value pairs that are incrementally revealed until the human expert reaches a satisfactory level of confidence. 
\textit{In contrast to the interactive explanations provided by xPACS, PROTEUS provides a global feature subspace that could potentially explain even unseen anomalies.}

\subsection{Explaining Black-box Predictors}
Several methods have been recently proposed to explain why a supervised model predicted a particular label for a particular sample \cite{FongV17,KohL17,MontavonSM18,Ribeiro0G16}. LIME \cite{Ribeiro0G16} constructs a linear interpretable model that is locally faithful to the predictor. In this respect, it draws uniformly at random (where the number of
such draws is also uniformly sampled) pseudo-samples per every sample to be explained. Note that LIME let the black-box classifier label the generated pseudo-samples. To the best of our knowledge, LIME has not been successfully used for imbalanced neighborhoods \cite{WhiteG20}. Other works \cite{FongV17,KohL17} explain the model by perturbing the features to quantify their influence on predictions. However, these works do not aim to explain multiple examples collectively, as the global explanation problem studied in our work.


Other works aim to produce explanations in the form of feature relevance scores, which indicate the relative importance of each feature to the classification decision. Such scores have been computed by comparing the difference between a classifier’s prediction score and the score when a feature is assumed to be unobserved \cite{RobnikSikonja2008ExplainingCF}, or by considering the local gradient of the classifier’s prediction score with respect to the features for a particular example \cite{BaehrensSHKHM10}. 
\cite{StrumbeljK10,StrumbeljK14} considered how to score features in a way that takes into account the joint influence of feature subsets on the classification score, which usually requires approximations due to the exponential number of such subsets.

The aforementioned works require as input a supervised model rather than an unsupervised anomaly detector. However, in real application settings it is difficult or even impossible to label data as anomalous or normal examples \cite{Goldstein2016}. Moreover, 
\textit{PROTEUS provides global explanations returned by standard feature selection algorithms after learning the decision boundary of the unsupervised detector.}

\subsection{Evaluation of Explainers}
Existing approaches for evaluating explanation methods in both supervised and unsupervised settings are typically quite limited in their scope. Often evaluations are limited to visualizations or illustrations of several example explanations \cite{BaehrensSHKHM10,DangANZS14} or to test whether a computed explanation collectively conforms to some known concept in the dataset \cite{BaehrensSHKHM10}, often for synthetically generated data. \cite{SiddiquiFDW19} proposes a larger scale quantitative evaluation methodology for anomaly explanations regarding sequential feature explanation methods. 
\textit{Compared to this study, in our work we assess the predictive performance of a classifier given an explanation along with the correctness of the learned features of the explanation.}

\subsection{Imbalanced Learning} 
One of the main challenges in supervised anomaly detection, is class imbalance: anomalies are largely underrepresented compared to normal examples. In the following we position PROTEUS w.r.t. the main imbalanced learning methods \cite{He2009LearningFI}. The imbalanced learning problem is concerned with the performance of learning algorithms in the presence of underrepresented data and severe class distribution skews. We follow the same categorization of imbalanced learning methods as in \cite{He2009LearningFI}.

\emph{Random oversampling} augments the original dataset by replicating examples from the minority class, while \emph{random undersampling} removes a random set of majority class examples. 
PROTEUS pipelines do not perform random under/over-sampling. The synthetic minority oversampling technique (SMOTE) \cite{Chawla2002SMOTESM} generates new minority class examples from the line segments that join the $k$ minority-class nearest neighbors.
Our pipeline generates synthetic examples close to the original minority examples by adding gaussian noise. SVM SMOTE \cite{Nguyen2011BorderlineOF} is a SMOTE variant that generates the synthetic examples concentrated in the most critical area, i.e., the boundary discovered by fitting an SVM classifier. Borderline-SMOTE \cite{Han2005BorderlineSMOTEAN} seeks to oversample the minority class instances in the borderline areas, by defining a set of “Danger” examples. Adaptive Synthetic Sampling (ADASYN) \cite{He2008ADASYNAS} algorithm uses a density distribution as a criterion to automatically decide the number of synthetic examples that need to be generated for each minority example. 
\textit{In comparison to the aforementioned works, PROTEUS performs a supervised synthetic minority oversampling ensuring that new samples are anomalies according to the decision boundary of an unsupervised detector that is currently explained. In addition, we proposed a method to avoid information leakage in the CV protocol when synthetic oversampling is applied.}


\section{Conclusion and Future Work}
\label{sec:conclusion}

In this paper we proposed the first methodology for producing predictive, global anomaly explanations in a detector-agnostic fashion. In particular, we show how with adequate design choices regarding rare class oversampling and unbiased performance estimation of ML pipelines, generating predictive, global anomaly explanations boils down to an AutoML problem. As yielded from our experiments, PROTEUS is not only able to discover explaining subspaces of features relevant to anomalies, but it can also construct predictive models that approximate effectively and robustly the decision boundary of popular unsupervised detectors (e.g., IF, LOF, LODA).
As future work, it would be interesting to approximate the decision boundary of a detector directly from the provided anomaly scores rather than converting them to binary labels. Hence, one could transform the explanation problem into regression with feature selection.


\begin{acks}
The research work was supported by the Hellenic Foundation for Research and Innovation (H.F.R.I.) under the “First Call for H.F.R.I. Research Projects to support Faculty members and Researchers and the procurement of high-cost research equipment grant” (Project Number: 1941); the ERC under the European Union's Seventh Framework Programme (FP/2007-2013) / ERC Grant  Agreement n. 617393.
\end{acks}


\bibliographystyle{ACM-Reference-Format}
\bibliography{biblio_short}

\end{document}